\documentclass[10pt,twocolumn,letterpaper]{article}

\usepackage[pagenumbers]{cvpr} 

\definecolor{vividcyan}{RGB}{0, 176, 240}
\definecolor{MyPurple}{HTML}{8E44AD}

\usepackage{booktabs}     
\usepackage{graphicx}     
\usepackage{xcolor}       
\usepackage{amssymb}    
\usepackage{pifont}     
\usepackage{colortbl}   
\usepackage{multirow} 
\usepackage{makecell}
\usepackage{bm}

\usepackage{booktabs}   
\usepackage{amssymb}    
\usepackage{pifont}     
\usepackage{colortbl}   
\usepackage{algorithmicx}

\usepackage{algorithm}       
\usepackage{algpseudocode}   

\usepackage[most]{tcolorbox}  
\usepackage{lipsum}         

\newtcolorbox{promptbox}[2][]{
  breakable,                
  width=\textwidth,         
  colback=black!5,          
  colframe=black!60,        
  arc=2mm,                  
  boxrule=0.8pt,            
  fonttitle=\bfseries,      
  title={#2},               
  #1                        
}

\definecolor{cvprblue}{rgb}{0.21,0.49,0.74}
\usepackage[pagebackref,breaklinks,colorlinks,allcolors=cvprblue]{hyperref}

\def\paperID{xxxx} 
\def\confName{---}
\def\confYear{---}

\title{LogicLens: Visual-Logical Co-Reasoning for Text-Centric Forgery Analysis
}

\author{
    Fanwei Zeng$^1$, Changtao Miao$^1$, Jing Huang$^1$, Zhiya Tan$^2$, Shutao Gong$^1$,\\
    Xiaoming Yu$^1$, Yang Wang$^1$, Huazhe Tan$^1$, Weibin Yao$^1$, Jianshu Li$^1$\\[2mm]
    $^1$Ant Group \qquad $^2$Nanyang Technological University\\
}

\begin{document}
\maketitle
\begin{abstract}
Sophisticated text-centric forgeries, fueled by rapid AIGC advancements, pose a significant threat to societal security and information authenticity.
Current methods for text-centric forgery analysis are often limited to coarse-grained visual analysis and lack the capacity for sophisticated reasoning. 
Moreover, they typically treat detection, grounding, and explanation as discrete sub-tasks, overlooking their intrinsic relationships for holistic performance enhancement.
To address these challenges, we introduce \textbf{LogicLens}, a unified framework for \textbf{Visual-Textual Co-reasoning} that reformulates these objectives into a joint task. 
The deep reasoning of LogicLens is powered by our novel \textbf{Cross-Cues-aware Chain of Thought (CCT)} mechanism, which iteratively cross-validates visual cues against textual logic. 
To ensure robust alignment across all tasks, we further propose a \textbf{weighted multi-task reward function} for GRPO-based optimization.
Complementing this framework, We first designed the \textbf{PR² (Perceiver, Reasoner, Reviewer)} pipeline, a hierarchical and iterative multi-agent system that generates high-quality, cognitively-aligned annotations. Then, we constructed \textbf{RealText}, a diverse dataset comprising \textbf{5,397} images with the fine-grained annotations, including textual explanations, pixel-level segmentation, and authenticity labels for model training.
Extensive experiments demonstrate the superiority of \textbf{LogicLens} across multiple benchmarks. 
In a zero-shot evaluation on T-IC13, it surpasses the specialized framework by \textbf{41.4\%} and GPT-4o by \textbf{23.4\%} in macro-average F1 score. 
\textbf{Moreover}, on the challenging dense-text T-SROIE dataset, it establishes a significant lead over other MLLM-based methods in mF1, CSS, and the macro-average F1. 
Our dataset, model, and code will be made publicly available.
\end{abstract}

\section{Introduction}
\label{sec:intro}

\begin{figure}[h!] 
  \centering
  \includegraphics[width=\columnwidth]{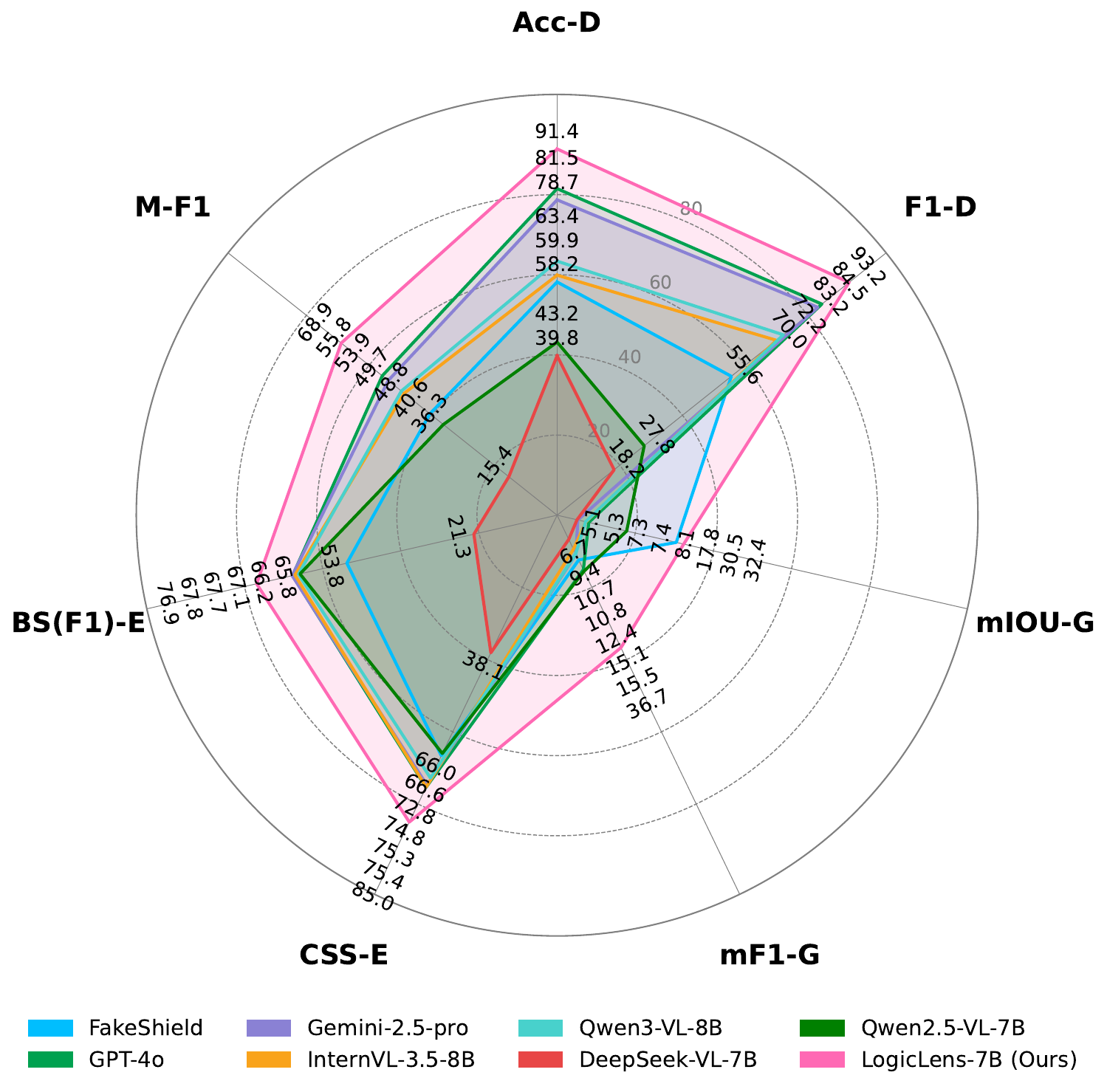}
  \caption{LogicLens(pink area) achieves \textbf{state-of-the-art} performance on our RealText benchmark across detection (D), grounding (G), and explanation (E), with M-F1 (macro-average F1 over the three tasks) as the unified evaluation metric.}
\label{fig:intro_radar_chart}
\end{figure}
The proliferation of generative AI, propelled by models like DALL-E~\cite{dalle3}, Midjourney~\cite{midjourney}, and Sora~\cite{liu2024sora,ramachandran2024sora}, is revolutionizing content creation. 
While these powerful tools~\cite{sauter2024flux,essentay2024stable,google2024gemini2.5}, unlock new frontiers of creativity.
they also enable the generation of increasingly sophisticated and visually indistinguishable forgeries, posing an unprecedented threat to information authenticity, particularly in the high-stakes domain of \textbf{text-centric forgery}. 
Attackers can manipulate textual content on critical carriers (e.g., official notices, financial receipts) to distort reality and inflict significant societal and economic harm. 
Developing robust methods to defend against such threats is therefore of critical importance for applications like media, judicial forensics and financial risk management.
\begin{figure*}[t]
  \centering
  \includegraphics[width=\textwidth, height=0.35\textwidth]{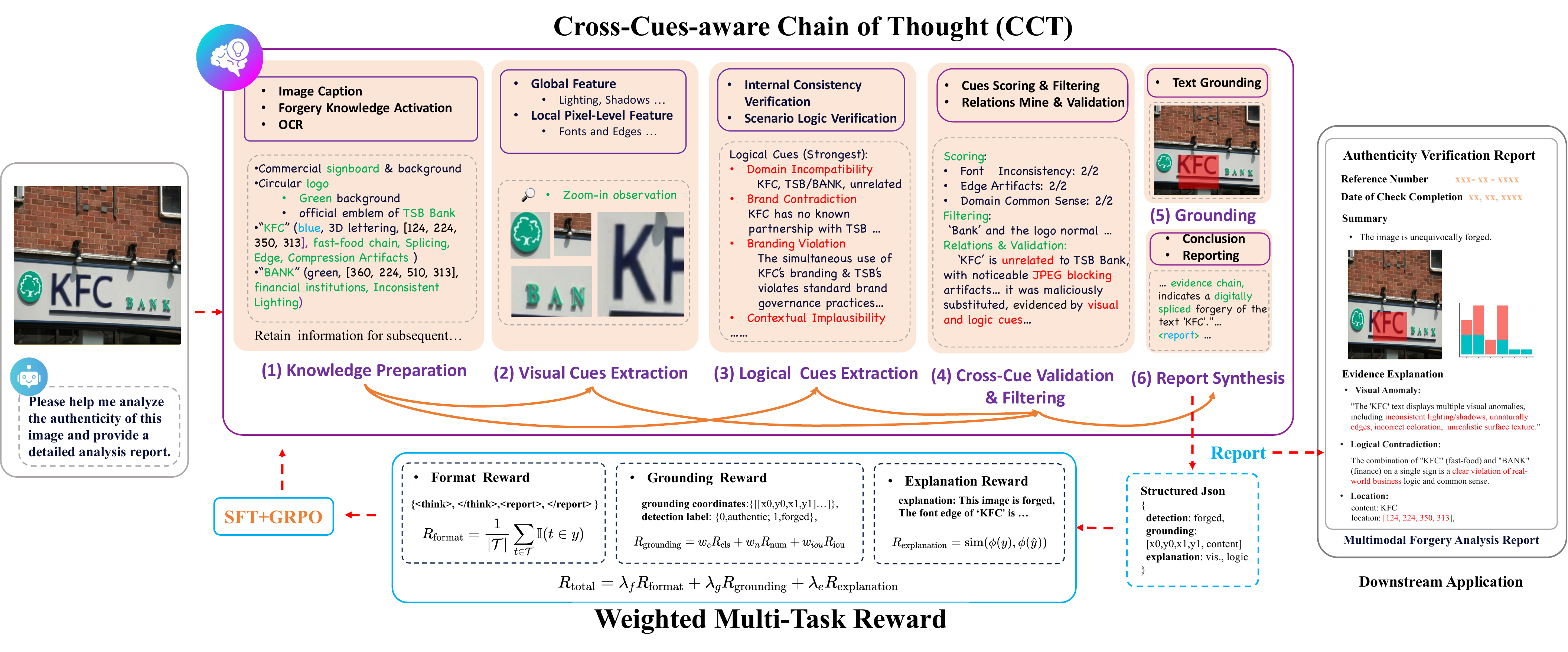} 
  \caption{Overview of the \textbf{LogicLens} framework. Our unified generative model takes an image and a prompt to autoregressively produce a structured textual forgery analysis~\textcolor{vividcyan}{\textit{\textbf{Report}}}. Central to this process is the \textbf{Cross-Cues-aware Chain of Thought (CCT)}, an iterative reasoning mechanism that analyzes and cross-validates visual and logical cues. The entire model is optimized via a GRPO-based \textbf{weighted multi-task reward}. The resulting machine-readable report is a versatile output that can power various downstream applications, such as the professional forensic report shown on the right.}
\label{fig:framework_overview} 
\end{figure*}
While traditional forensic methods often focus on coarse, image-level classification and inherently lack explainability~\cite{lampert2006printing, chen2024single,duan2024test,jeong2022frepgan,tan2024frequency,wang2023dynamic, wu2019mantra,kwon2022learning,hu2020span,cozzolino2020noiseprint,chen2021image}, recent MLLM-based works have started to address the need for explainable analysis~\cite{qu2024textsleuth,xu2025fakeshield,gao2025fakereasoning,kang2026legion,chang2024antifakeprompt}. 
However, they remain limited by a fragmented architecture that treats detection, grounding, and explanation as separate tasks, hindering synergistic reasoning and holistic performance enhancement. Additionally, their analysis often overrelies on visual artifacts while neglecting critical logical inconsistencies in the text. This decoupled design and dependence on shallow cues reduce robustness, especially when visual traces are subtle or ambiguous.
Furthermore, as regulatory and societal demands for transparency intensify, there is a compelling need for a more explainable paradigm in forgery analysis~\cite{karatsiolis2024survey,nist2023ai,a&o2024euaiact}. 
Such a paradigm must provide not only a forgery score but also the precise location of manipulations and a reasoned analysis of key forgery clues, as illustrated in Fig.~\ref{fig:framework_overview}. We argue that a deep, cross-modal reasoning process that fuses both visual and logical cues is essential for robust explainable analysis.

To address these challenges, we introduce \textbf{LogicLens}, a unified framework that reformulates the three objectives into a holistic generative paradigm. 
The deep reasoning of LogicLens is powered by our novel \textbf{Cross-Cue aware Chain-of-Thought (CCT)} mechanism, which comprises 6 interdependent stages: (1) Knowledge Preparation, (2) Visual Cue Extraction, (3) Logical Cue Extraction, (4) Cross-Cue Validation \& Filtering, (5) Grounding, and (6) Report Synthesis. 
The collaboration of these stages enables the model to not only identify disparate visual and logical clues but also perform a critical cross-verification of this evidence. 
To further enhance performance and stability, we introduce a specialized \textbf{weighted multi-task reward function}. 
This function, which balances distinct components for format adherence, grounding accuracy, and explanation quality, is optimized with a combination of SFT and GRPO to robustly align the model with all task objectives.
Furthermore, we construct \textbf{RealText}, a new benchmark specifically designed for deep, text-centric forgery analysis and model training. 
Its high-quality annotations were generated via our novel \textbf{PR²} pipeline in Fig.~\ref{fig:data_pipeline}, a hierarchical and self-correcting multi-agent system. 
This pipeline simulates a forensic workflow where a \textbf{Perceiver} agent forms an initial analysis draft, a \textbf{Reasoner} generates structured rationale, and a \textbf{Reviewer} performs quality control and provides feedback for iterative refinement.
Our main contributions are summarized as follows:
\begin{itemize}
    \item We propose LogicLens, a unified generative framework that reformulates detection, grounding, and explanation into a single, joint task. The capability of deep reasoning is unlocked by our novel CCT mechanism with visual and textual clues cross-validation, and optimization via our GRPO-guided multi-task reward function.

    \item We introduce RealText, the first large-scale benchmark for explainable text-centric forgery analysis that includes annotations for textual logical clues. Its fine-grained, multi-task annotations were generated via our novel PR² pipeline, a hierarchical and self-correcting multi-agent system.

    \item Extensive experiments show that LogicLens sets a new state-of-the-art on our benchmark and demonstrates superior zero-shot generalization and robustness on existing datasets.
\end{itemize}


\section{Related Works}
\label{sec:formatting}

\subsection{Forgery Analysis in Text-Centric Images}
\label{subsec:forgery_detection}
Conventionally, the analysis of manipulated text in images has been framed as “Tampered Text Detection”~\cite{ahmed2014forgery,wang2022tampered,song2024cross,bertrand2015conditional}, focusing primarily on binary forgery classification of textual regions. More recently, a shift toward interpretability has given rise to “Explainable Tampered Text Detection”~\cite{qu2024textsleuth}, which introduces explanatory capabilities but often compromises robust detection and fine-grained grounding in favor of interpretability.
Although early methods were not designed for comprehensive analysis, their forensic principles offer valuable insights for developing more advanced, reasoning-based approaches. For instance, traditional models such as DTD~\cite{qu2023towards} and CAFTB-Net~\cite{song2024cross} effectively leverage visual and frequency-domain artifacts—such as boundary inconsistencies and block distortions—for detection, while TextSleuth~\cite{qu2024textsleuth} demonstrates the potential of MLLMs in generating explanations based on visual cues. Nevertheless, these approaches mostly rely on visual signals and largely neglect logical contradictions embedded within the textual content itself. 

\subsection{Explainable Forgery Analysis with MLLMs}
\label{subsec:explainable_analysis}

The emergence of MLLMs has advanced explainable forensic systems along two main lines. The first, prompt-driven analysis, leverages pre-trained MLLMs via zero-shot inference (e.g., GPT-4o) or prompt tuning (e.g., AntifakePrompt~\cite{chang2024antifakeprompt}) but is limited to image-level classification and lacks fine-grained grounding. The second stream employs segmentation-based architectures like FakeShield~\cite{xu2025fakeshield} and LEGION~\cite{kang2026legion}, yet relies on decoupled, multi-stage pipelines that first generate coarse textual descriptions and then produce segmentation masks in separate modules. This cascaded design suffers from error propagation and impedes cross-modal reasoning. Although LEGION~\cite{kang2026legion} explored physical law violations in images, it is biased toward fully synthetic natural images. Most approaches~\cite{lai2024lisa,rasheed2024glamm,huang2024sida} prioritize visual cues while neglecting logical and semantic contradictions in the text, highlighting the need for a unified, end-to-end framework capable of deep cross-modal forgery analysis.

\section{Methodology}
\label{sec:methodology}

Inspired by recent advancements in MLLM-based forgery analysis~\cite{lai2024lisa, xu2025fakeshield,kang2026legion,nguyen2024editscout,li2024forgerygpt,sun2024forgerysleuth,gao2025fakereasoning}, we introduce LogicLens, a unified reasoning framework designed to address the limitations discussed in Section~\ref{sec:intro}. The foundation of LogicLens is its end-to-end architecture, which integrates detection, grounding, and explanation into a single, cohesive process. This unified design directly addresses the fragmentation in prior work, allowing for synergistic reasoning that improves overall performance.
Building upon this architecture, we introduce a novel Cross-Cues-aware Chain of Thought (CCT) mechanism. 
This core component moves beyond analyzing clues in isolation, empowering LogicLens to fuse evidence from both visual artifacts and logical reasoning to form a more holistic and reliable judgment.
To ensure these complementary capabilities are optimized jointly, we employ a weighted multi-task reward function trained with the GRPO algorithm, promoting both stable and high-performing outputs.

\subsection{Unified Generative Formulation}
\label{subsec:overview}

In this section, we define~\textbf{text-centric forgery analysis}, a unified challenge that requires models to simultaneously detect, ground, and explain manipulations in any image containing text.
To achieve this, we reformulate the entire challenge into a single, joint generative task, eliminating the need for separate, task-specific decoders. 
Specifically, the model learns to autoregressively generate a single textual analysis report, denoted as \(R_{\text{analysis}}\), that encompasses all three outputs: a textual classification (e.g., "HIGH INDICATION OF FORGERY")for detection (\(S_{\text{det}}\)), a coordinate string (e.g., "[x1, y1, ...x4, y4]") for grounding (\(L_{\text{coords}}\)), and a detailed rationale for explanation (\(E_{\text{rationale}}\)). 
This unified process, where the entire report is generated by a single MLLM decoder \(\mathcal{D}\) conditioned on the input image \(I\) and task prompt \(T\) processed by an encoder \(\mathcal{E}\), can be expressed as:
\begin{equation}
    R_{\text{analysis}} \leftarrow \mathcal{D}(\mathcal{E}(I, T))
    \label{eq:unified_generation}
\end{equation}. 
The structured output also enables numerous downstream applications, such as automatically generating professional, multi-modal forensic documents in Fig.~\ref{fig:framework_overview}. 

\subsection{Cross-Cue Thought Chain (CCT)}
\label{subsec:cct}

To address the challenge of effectively analyzing and validating both visual and logical anomaly cues, we introduce the CCT. This specialized reasoning mechanism is engineered to correlate information across these two modalities, thereby significantly enhancing the model's deep reasoning capabilities for forgery analysis.
As illustrated in Figure~\ref{fig:framework_overview}, this cognitive workflow consists of the following key stages, the entire process is detailed in \cref{alg:LogicLens}:

\noindent\textbf{Knowledge Preparation.}
This initial stage aims to establish a holistic understanding of the image, from its scene to textual content, and to activate relevant forensic knowledge, thereby preparing a foundation for subsequent cues extraction, as illustrated by the green text in Fig.~\ref{fig:framework_overview}~(1). This process internally consists of several key steps:
First, the model performs an implicit image captioning to grasp the overall context, 
This contextual understanding is then used to activate its internal domain-specific knowledge. The model identifies key textual entities from the caption (e.g., "KFC," "BANK," "Logo") and retrieves associated world knowledge (e.g., "fast-food chain"). 
Concurrently, its internal forensic knowledge is also activated, prompting the model to consider potential manipulation types relevant to the scene (e.g., copy-move, text editing, AIGC inpainting). 
Finally, building upon this preparatory work, a OCR process is executed in parallel to extract the precise coordinates and structural information of all text.

\noindent\textbf{Visual Cues Extraction.}
The goal of this stage is to identify visual anomaly cues by performing both global and local perceptual analysis. 
As shown in Fig.~\ref{fig:framework_overview} (2),the model first assesses global characteristics such as lighting, shadows. 
It cross-references these observations with its understanding of physical imaging principles to detect inconsistencies (e.g., "overhead spotlight light with similar shadows"). 
Subsequently, the model performs a local, pixel-level inspection of suspected high-risk regions. 
This process, analogous to an analyst zooming in on an image in Fig.~\ref{fig:framework_overview}~(2), scrutinizes fine-grained artifacts(e.g., edges, font styles, color consistency). 
Finally, all identified visual cues are aggregated to form a preliminary conclusion based on perceptual evidence.

\begin{algorithm}[t]
\caption{Cross-Cue aware Chain of Thought (CCT) Process}
\label{alg:LogicLens}
\begin{algorithmic}[1]
\Statex \textbf{Input:} Image $I$
\Statex \textbf{Output:} Structured Analysis Report $R$
\Statex 

\Statex \textbf{Stage 1: }
\State $C_{\text{caption}} \gets \text{ImplicitImageCaptioning}(I)$
\State $K_{\text{world}} \gets \text{ActivateWorldKnowledge}(C_{\text{caption}})$
\State $K_{\text{forensic}} \gets \text{ActivateForensicKnowledge}(C_{\text{caption}})$
\State $D_{\text{ocr}} \gets \text{ExtractTextInfo}(I)$

\Statex \textbf{Stage 2: }
\State $Cues_{\text{global}} \gets \text{AnalyzeGlobalConsistency}(I, K_{\text{forensic}})$
\State $Cues_{\text{local}} \gets \text{InspectLocalAnomalies}(I, K_{\text{forensic}})$
\State $V_{\text{cues}} \gets \text{AggregateCues}(Cues_{\text{global}}, Cues_{\text{local}})$

\Statex \textbf{Stage 3:}
\State $Cues_{\text{internal}} \gets \text{VerifyInternalConsistency}(D_{\text{ocr}})$
\State $Cues_{\text{scenario}} \gets \text{VerifyScenarioLogic}(D_{\text{ocr}}, K_{\text{world}})$
\State $L_{\text{cues}} \gets \text{AggregateCues}(Cues_{\text{internal}}, Cues_{\text{scenario}})$

\Statex \textbf{Stage 4: }
\State $All_{\text{cues}} \gets V_{\text{cues}} \cup L_{\text{cues}}$
\State $Scores \gets \text{InternalScoringMechanism}(All_{\text{cues}})$
\State $HighValue_{\text{cues}} \gets \text{FilterByScore}(All_{\text{cues}}, Scores)$

\Statex \textbf{Stage 5:}
\State $TamperedRegions \gets \emptyset$
\ForAll{$cue \in HighValue_{\text{cues}}$}
    \State $matched\_text \gets \text{SemanticRetrieval}(cue, D_{\text{ocr}})$
    \State $TamperedRegions \gets TamperedRegions \cup \text{GetCoordinates}(matched\_text)$
\EndFor

\Statex \textbf{Stage 6:}
\State $Verdict \gets \text{DetermineFinalVerdict}(HighValue_{\text{cues}})$
\State $Rationale \gets \text{ConsolidateEvidence}(HighValue_{\text{cues}})$
\State $R \gets \text{GenerateReport}(Verdict, Rationale,$
\Statex \hspace{\algorithmicindent} $TamperedRegions)$

\State \textbf{return} $R$
\end{algorithmic}
\end{algorithm}
\noindent\textbf{Logical Cues Extraction.}
This stage goes beyond visual analysis and performs a higher-level semantic check, leveraging the OCR results from Stage 1 and the MLLM’s built-in world knowledge, as shown in Fig.~\ref{fig:framework_overview}~(3).

To handle the diversity of document types, the model is guided to identify logical contradictions along two primary dimensions:

\begin{itemize}
    \item \textbf{Internal Consistency Verification:} This involves evaluating the document's internal consistency by verifying quantifiable elements, such as \textit{arithmetical calculations} (e.g., \texttt{$2048.3 - 1024.1 \neq 1023.1$}) and checking for temporal paradoxes or other sequential impossibilities.
    
    \item \textbf{Scenario Logic Verification:} This assesses the document's external plausibility. The model cross-references its content against real-world knowledge and common sense to detect contextual anomalies (e.g., "The juxtaposition of 'KFC', 'BANK', a severe business logic conflict").
\end{itemize}

All identified logical inconsistencies are then aggregated to form a set of high-level semantic cues for the subsequent cross-validation stage.

\noindent\textbf{Cross-Cues Validation \& Filtering.}
This stage cross-validates the previously identified visual and logical cues, removes redundant findings, and enhances the overall evidential value. To this end, LogicLens employs a specialized scoring mechanism that evaluates the significance of each cue. Based on these scores, only the most valuable and robust anomalies are retained for the final conclusion. Notably, this scoring mechanism is internalized within the model during training, eliminating the need for an external, rule-based ``scoring prompt'' and thereby avoiding its inherent rigidity.

\noindent\textbf{Grounding.}
The objective of this stage is to precisely localize tampered regions by providing both their coordinates and text. 
Building on the high-value clues retained from the cross-validation stage, the model performs targeted localization by reactivating the fine-grained OCR output from Stage 1. 
It then matches these validated forgery indicators against the structured OCR data through a semantic retrieval process, pinpointing and outputting the exact coordinates of the manipulated text, as visualized in Fig.~\ref{fig:framework_overview}~(5).

\noindent\textbf{Report Synthesis.}
In this final stage, all validated findings from the preceding stages, the detection verdict, grounding coordinates, and supporting rationale, are internally consolidated. The generation of the structured analysis report is then triggered by a special \texttt{<report>} token, prompting the model to render these consolidated evidences into a coherent, machine-readable format.

\subsection{Weighted Multi-Task Reward Function.}
\label{subsec:reward_function}
Our composite reward function \(R_{\text{total}}\) is a weighted sum of three distinct components, inspired by~\cite{huang2025lavcot} each targeting a critical aspect of the final output:
\begin{equation}
    R_{\text{total}} = \lambda_f R_{\text{format}} + \lambda_g R_{\text{grounding}} + \lambda_e R_{\text{explanation}}.
\end{equation}
Let \(y\) be the model's generated report and \(\hat{y}\) be the ground-truth report. 
In our experiments, these weights are empirically set to \(\lambda_f = 0.15\), \(\lambda_g = 0.75\), and \(\lambda_e = 0.1\).
Each component is detailed as follows:
\begin{itemize}
    \item \textbf{Format Adherence Reward (\(R_{\text{format}}\)):} To encourage well-structured outputs, this reward is based on the presence of key structural tags. Let \(\mathcal{T} = \{\text{\texttt{<think>}}, \text{\texttt{</think>}}, \text{\texttt{<report>}}, \text{\texttt{</report>}}\}\) be the set of required tags. The reward is the proportion of tags present in the generated output \(y\):
    \begin{equation}
        R_{\text{format}} = \frac{1}{|\mathcal{T}|} \sum_{t \in \mathcal{T}} \mathbb{I}(t \in y),
    \end{equation}
    where \(\mathbb{I}(\cdot)\) is the indicator function.

    \item \textbf{Grounding Accuracy Reward (\(R_{\text{grounding}}\)):} This is a multi-faceted reward evaluating detection and localization. Let \(c\) and \(\hat{c}\) be the predicted and ground-truth binary forgery labels, respectively. Let \(B\) and \(\hat{B}\) be the set of predicted and ground-truth bounding boxes. The reward is an aggregation of three sub-rewards:
    \begin{equation}
        R_{\text{grounding}} = w_c R_{\text{cls}} + w_n R_{\text{num}} + w_{iou} R_{\text{iou}}.
    \end{equation}
    \begin{enumerate}
        \item \textit{Detection Accuracy (\(R_{\text{cls}}\)):} A binary reward for correct classification:
        \begin{equation}
            R_{\text{cls}} = 1.0 \cdot \mathbb{I}(c = \hat{c}).
        \end{equation}
        \item \textit{Object Count Precision (\(R_{\text{num}}\)):} This reward encourages predicting the correct number of tampered regions. It is defined as:
        \begin{equation}
            R_{\text{num}} = 
            \begin{cases} 
                0.5 \cdot \mathbb{I}(|B| = |\hat{B}|) & \text{if } \hat{c} = \text{forged} \\
                0.5 \cdot \mathbb{I}(|\hat{B}| = 0)   & \text{if } \hat{c} = \text{authentic}
            \end{cases}
        \end{equation}
        \item \textit{Localization IoU Score (\(R_{\text{iou}}\)):} 
            This reward evaluates spatial accuracy based on the mean Intersection over Union (mIoU). 
            First, for each predicted box \(\hat{b} \in \hat{B}\), we find its maximum IoU with any ground-truth box \(b \in B\). We then compute the mIoU by averaging these scores:
            \begin{equation}
                \text{mIoU} = \frac{1}{|\hat{B}|} \sum_{\hat{b} \in \hat{B}} \max_{b \in B} \text{IoU}(\hat{b}, b)
                \label{eq:miou_calc}
            \end{equation}
            This mIoU score is then mapped to a discrete reward value, \(R_{\text{iou}}\), to incentivize high-quality localization:
            \begin{equation}
                R_{\text{iou}} =
                \begin{cases}
                    0.6 & \text{if mIoU} > 0.8 \\
                    0.4 & \text{if } 0.5 \le \text{mIoU} \le 0.8 \\
                    0.0 & \text{otherwise}
                \end{cases}
                \label{eq:riou_tiered}
            \end{equation}
    \end{enumerate}
    
    \item \textbf{Explanation Quality Reward (\(R_{\text{explanation}}\)):} To evaluate the semantic quality of the generated explanation, we compute the cosine similarity between the sentence embeddings of the generated text \(y\) and the ground-truth \(\hat{y}\). We use a pre-trained sentence embedding model, \(\phi(\cdot)\) (Qwen3-embedding-4B). The cosine similarity, \(\text{sim}(\cdot, \cdot)\) measured their semantic alignment:
    \begin{equation}
        R_{\text{explanation}} = \text{sim}(\phi(y), \phi(\hat{y}))
    \end{equation}
\end{itemize}

\section{Datasets}
\label{sec:dataset}
\subsection{Motivation}
\label{subsec:motivation_principles}
\begin{figure*}[t]
  \centering
  \includegraphics[width=\textwidth, height=0.35\textwidth]{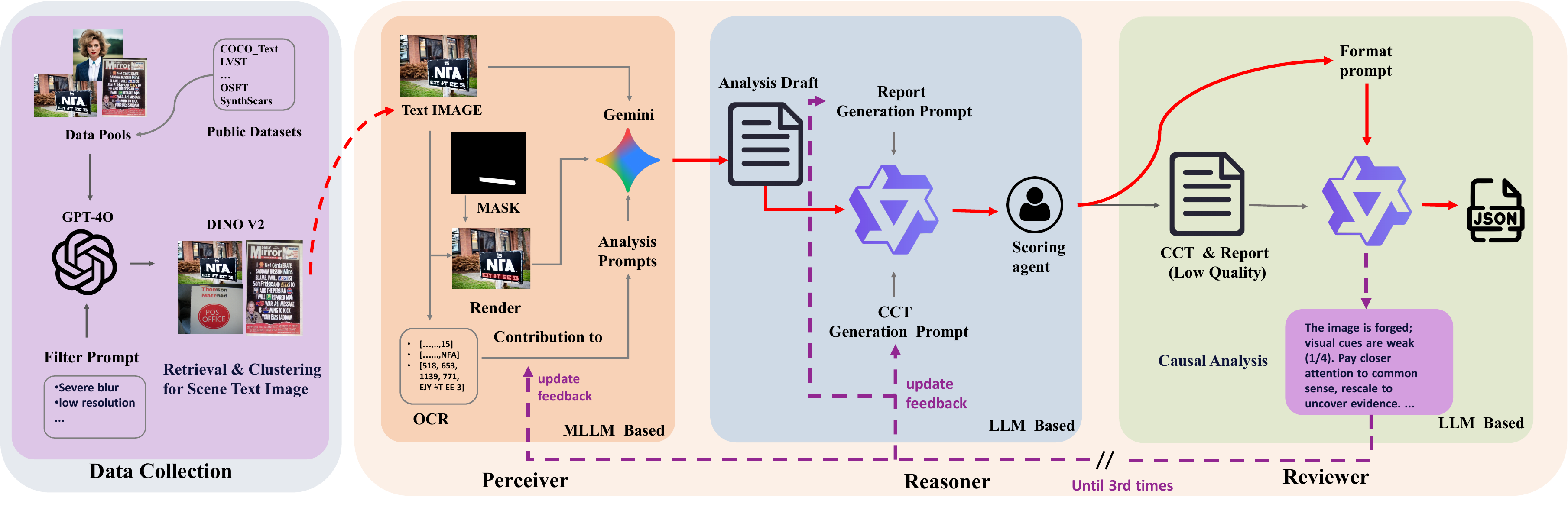}
  \caption{The architecture of our \textbf{PR²} (\textbf{P}erceiver, \textbf{R}easoner, \textbf{R}eviewer) pipeline. After an initial data collection stage, our multi-agent system generates annotations through a collaborative, iterative process. The Perceiver drafts an analysis, the Reasoner structures it to target CCT \& analysis report, and the Reviewer validates its quality, initiating a \textbf{refinement loop} if necessary. 
  This cycle, indicated by the solid~\textcolor{red}{$\bm{\rightarrow}$} and dashed~\textcolor{MyPurple}{$\bm{\dashleftarrow}$} feedback loops,
  ensures the final output is a high-fidelity, structured JSON annotation.}
  \label{fig:data_pipeline}
\end{figure*}


A comprehensive benchmark for text-centric forgery analysis is notably absent. While existing datasets have made valuable contributions, each addresses only a fragment of the problem. T-IC13~\cite{wang2022detecting} and T-SROIE~\cite{wang2022tampered} provide foundational forged samples but are limited by scale and imbalance. DocTamper~\cite{qu2023towards} offers large-scale data but discards global semantic context and lacks modern AIGC forgery types. OSFT~\cite{qu2024revisiting} enhances data diversity with diffusion models. However, none of these benchmarks provide the fine-grained, explanatory annotations required for deep reasoning. Even the most recent proposal, ETTD~\cite{qu2024textsleuth}, maintains a visual-only focus and is not yet publicly available.
\subsection{Data Curation Pipeline}
\label{subsec:data_curation}
\begin{table}[t]
  \centering
  \caption{RealText is the first public benchmark to provide comprehensive annotations for text-centric forgery analysis tasks: detection (Det), grounding (Mask), and explanation (Expl).}
  \label{tab:dataset_comparison_final_complete}

  \setlength{\tabcolsep}{3.5pt} 

  \begin{tabular}{lccccc}
    \toprule
    \textbf{Dataset} & \textbf{Total} & \textbf{Textline} & \textbf{\makecell{Det}}  & \textbf{Mask}  & \textbf{Expl}\\
    \midrule
    T-IC13~\cite{wang2022detecting}      & 462       & \checkmark & \ding{55} & \ding{55} & \ding{55} \\
    T-SROIE~\cite{wang2022tampered}     & 986       & \checkmark & \ding{55} & \ding{55} & \ding{55} \\
    OSFT~\cite{qu2024revisiting}        & 2938      & \checkmark & \ding{55} & \ding{55} & \ding{55} \\
    DocTamper~\cite{qu2023towards}   & 170,000   & \ding{55}  & \ding{55} & \checkmark & \ding{55} \\
    \midrule
    \textbf{RealText (Ours)} & \textbf{5397} & \checkmark & \checkmark & \checkmark & \checkmark \\
    \bottomrule
  \end{tabular}
\end{table}
To address these limitations, we developed the \textbf{PR²} pipeline, a hierarchical and feedback-driven multi-agent system for data curation, as illustrated in Fig.~\ref{fig:data_pipeline}. 
Powered by it, we constructed our comprehensive dataset, \textbf{RealText}.

To ensure a diverse foundation for PR², we first aggregated samples from various public benchmarks~\cite{veit2016cocotext,zhang2019icdarrects,sun2019icdar,kang2026legion,qu2024revisiting} and then applied a rigorous filtering process with GPT-4o~\cite{openai2024gpt4o} and DINOv2~\cite{oquab2023dinov2} to yield a collection of text-centric images. 
Based on these images, the corresponding fused mask-image and the OCR transcript are then prepared to initiate the PR².This input subsequently flows through our three specialized agents:
\begin{itemize}
    \item \textbf{The Perceiver: Holistic Forgery Analysis.} 
    Guided by a carefully designed analysis prompt (detailed in Appendix \textbf{B}), the Perceiver agent, powered by Gemini-2.5-pro~\cite{google2024gemini2.5}, performs a comprehensive assessment of the highlighted region on the fused mask-image. It then generates a preliminary report detailing both visual and logical anomaly cues. This initial draft, while less structured than the final CCT, provides the raw material for refinement.

    \item \textbf{The Reasoner: Structured Refinement and Scoring.} The Reasoner (Qwen3-Max) takes the Perceiver's raw text as input. Guided by specialized prompts, it performs two sequential tasks: first, it structures the raw analysis into a formalized CCT and a coherent report; second, it assigns a quality score based on global dimensions such as format integrity, conclusion consistency, and logical completeness.
    
    \item \textbf{The Reviewer: Iterative Verification and Refinement.} This Reviewer acts as the final judge of quality and orchestrates the refinement loop. It first evaluates the annotation against a predefined quality threshold (e.g., 98/100). High-scoring annotations are accepted and passed to the final formatting stage. For low-quality samples, the Reviewer performs an automated cause analysis to generate constructive feedback. This feedback is then provided to the Perceiver and Reasoner, initiating an iterative refinement cycle. The cycle for each sample iterates up to a maximum of three times. Finally, all successfully validated annotations are structured into a JSON format to facilitate the multi-task reward calculation described in Section~\ref{subsec:reward_function}.
    
\end{itemize}
Leveraging the synergistic and iterative design of the PR² pipeline, the high-fidelity and cognitively-aligned annotations, generated automatically at scale. This provides a foundational resource for the development of LogicLens.

\subsection{Dataset Statistics and Comparison}
\label{subsec:dataset_stats}
As detailed in Table~\ref{tab:dataset_comparison_final_complete}, we present RealText, a new benchmark of 5,397 images designed to advance research in text-centric forgery analysis. Its key distinction is the unparalleled comprehensiveness of fine-grained annotations, which unify detection, grounding, and explanation into a single framework. Moreover, RealText features a challenging subset of 905 dense-text images, specifically included to facilitate rigorous evaluation of model robustness. As such, RealText represents a vital new resource for the community.


\section{Experiments}
\label{sec:experiments}
\subsection{Experimental Setup}
\label{subsec:setup}
\noindent\textbf{Implementation Details.}
Our framework is initialized with Qwen2.5-VL-7B~\cite{Qwen2.5-VL}. 
The training process consists of two stages: (1) Supervised Fine-Tuning (SFT) for 30 epochs, where we use LoRA~\cite{hu2022lora} (\(r=32, \alpha=64\)) on all attention projection layers with a learning rate of \(1 \times 10^{-4}\); and (2) a subsequent GRPO-based reinforcement learning stage for 10 epochs with a lower learning rate of \(1 \times 10^{-5}\). 
We resize all input images to \(896 \times 1344\) and add \texttt{<think>} and \texttt{<report>} as special tokens to the vocabulary. 
All experiments were conducted on 8 NVIDIA A100 GPUs.
our main training and evaluation are conducted on the RealText benchmark. Additionally, the model's out-of-domain generalization to sparse scene text is evaluated in a zero-shot setting on the T-IC13 dataset. Separately, we perform an independent experiment on the challenging, dense-text T-SROIE benchmark to investigate the model's robustness.
\begin{table*}[t]
  \centering
  \caption{Comprehensive performance comparison across three benchmarks: proposed RealText, T-IC13 (\textbf{zero-shot}) and T-SROIE (\textbf{dense text for robustness}), LogicLens demonstrates superior performance, strong generalization and robustness. All metrics are in percentage (\%). Acc: Accuracy, mIOU: mean Intersection over Union, CSS: Cosine Similarity Score, BS(F1): BERTScore F1. M-F1: the macro-average of its Detection F1, Grounding mF1, and Explanation BS(F1). Best results are in \textbf{bold}, second best are \underline{underlined}, '-': Model unreleased and result not reported.}
  \label{tab:main_results}
  \resizebox{\textwidth}{!}{%
    \begin{tabular}{@{}l|cc|cc|cc|c|cc|cc|cc|c|cc|cc|cc|c@{}}
      \toprule
      \multirow{3}{*}{\textbf{Method}} & \multicolumn{7}{c|}{\textbf{RealText}} & \multicolumn{7}{c|}{\textbf{T-IC13}} & \multicolumn{7}{c}{\textbf{T-SROIE}} \\
      \cmidrule(l){2-8} \cmidrule(l){9-15} \cmidrule(l){16-22}
      
       & \multicolumn{2}{c|}{Detection} & \multicolumn{2}{c|}{Grounding} & \multicolumn{2}{c|}{Explanation} & \multirow{2}{*}{\shortstack{M-F1}}
       & \multicolumn{2}{c|}{Detection} & \multicolumn{2}{c|}{Grounding} & \multicolumn{2}{c|}{Explanation} & \multirow{2}{*}{\shortstack{M-F1}}
       & \multicolumn{2}{c|}{Detection} & \multicolumn{2}{c|}{Grounding} & \multicolumn{2}{c|}{Explanation} & \multirow{2}{*}{\shortstack{M-F1}} \\
      \cmidrule(l){2-3} \cmidrule(l){4-5} \cmidrule(l){6-7} 
      \cmidrule(l){9-10} \cmidrule(l){11-12} \cmidrule(l){13-14} 
      \cmidrule(l){16-17} \cmidrule(l){18-19} \cmidrule(l){20-21} 
      
       & Acc & F1 & mIOU & mF1 & CSS & BS(F1) & 
       & Acc & F1 & mIOU & mF1 & CSS & BS(F1) & 
       & Acc & F1 & mIOU & mF1 & CSS & BS(F1) & \\ 
       \midrule
       
      \multicolumn{21}{l}{\textit{Group 1: State-of-the-Art MLLM-based Frameworks}} \\
      FakeShield~\cite{xu2025fakeshield} & 58.2 & 55.6 & \underline{30.5} & 12.4 & 66.6 & 53.8 & 40.6 & 36.9 & 32.3 & \underline{23.8} & \underline{26.2} & 57.5 & 52.7 & 37.1 & 14.7 & 25.7 & 0.5 & \underline{1.9} & 51.7 & 52.9 & 26.8  \\
      TextSleuth-7B~\cite{qu2024textsleuth} & - & - & - & - & - & - & - & \underline{88.4} & - & - & - & - & - & - & - & - & - & - & - & - & -  \\
      \midrule
      \multicolumn{21}{l}{\textit{Group 2: Powerful General-Purpose MLLMs}} \\
      gpt-4o-o3-0416~\cite{openai2024gpt4o} & \underline{81.5} & \underline{84.5} & 8.1 & 15.1 & \underline{75.4} & 67.7 & \underline{55.8} & 85.7 & \underline{90.1} & 22.9 & 13.5 & 69.1& 65.7 & \underline{56.4} & 84.5 & 91.6 & \underline{1.3} & 1.3 & 73.3 & 65.5 & 52.8  \\
      Gemini-2.5-Pro~\cite{google2024gemini2.5} & 78.7 & 83.2 & 5.3 & 10.8 & 74.8 & \underline{67.8}  & 53.9 & 80.4 & 87.0 & 17.9 & 12.1 & \underline{76.9} & \underline{68.7}  & 55.9 & \textbf{99.9} & \textbf{99.9} & 0.2 & 1.2 & \underline{79.8} & \underline{69.8} & \underline{57.0}  \\
      InternVL-3.5-8B~\cite{wang2025internvl3.5} & 59.9 & 70.0 & 7.4 & 9.4 & 75.3 & 67.1 & 48.8 & 63.1 & 73.5 & 15.9 & 9.8 & 71.4 & 66.4 & 49.9 & 86.4 & 92.7 & 0.1 & 1.0 & 74.7 & 67.5 & 53.7  \\
      qwen3-vl-8B~\cite{qwen3technicalreport} & 63.4 & 72.2 & 7.3 & 10.7 & 72.8 & 66.2 & 49.7 & 80.3 & 88.0 & 12.7 & 14.5 & 72.1 & 66.5  & 56.3 & \textbf{99.9} & \textbf{99.9} & 0.0 & 1.0 & 75.9 & 66.0 & 55.6  \\
      DeepSeekVL-7B~\cite{deepseek2024deepseekvl} & 39.8 & 18.2 & 5.1 & 6.7 & 38.1 & 21.3 & 15.4 & 39.1 & 35.5 & 5.0 & 11.3 & 37.2 & 28.9 & 25.2 & 11.3 & 20.3 & 0.0 & 0.5 & 60.6 & 53.7 & 24.8 \\
      Qwen2.5-VL-7B~\cite{Qwen2.5-VL} & 43.2 & 27.8 & 17.8 & \underline{15.5} & 66.0 & 65.8 & 36.3 & 38.2 & 38.5 & 15.1 & 9.0 & 64.4 & 65.5 & 37.7 & 6.1 & 11.5 & 0.0 & 0.7 & 65.4 & 66.3  & 26.2  \\
      \midrule
      \rowcolor{gray!15}
      \textbf{LogicLens (Ours)} & \textbf{91.4} & \textbf{93.2} & \textbf{32.4} & \textbf{36.7} & \textbf{85.0}&\textbf{76.9} & \textbf{68.9} & \textbf{91.2} & \textbf{93.2} & \textbf{44.5} & \textbf{67.6} & \textbf{86.7}&\textbf{78.5} & \textbf{79.8} & \underline{98.8} & \underline{99.4} & \textbf{9.1} & \textbf{11.0} & \textbf{84.9}&\textbf{77.0} & \textbf{62.5}  \\
      \bottomrule
    \end{tabular}%
  }
\end{table*}
\noindent\textbf{Evaluation Metrics.}
Our evaluation protocol employs a comprehensive suite of metrics for a holistic assessment. Performance on the detection task is measured using standard classification metrics, including Accuracy and F1-Score. For grounding, we assess localization accuracy via pixel-level mean Intersection over Union (mIoU) and mean F1-Score (mF1), adhering to the TruFor protocol~\cite{guillaro2023trufor}. The quality of the generated explanation is quantified by its semantic fidelity, using the Cosine Similarity Score (CSS) and BERTScore~\cite{zhang2020bertscore}. 
Finally, to provide a single, comprehensive indicator of overall performance on the text-centric forgery analysis task, we compute the Macro-Average F1-Score (M-F1) across all three components.
\begin{table}[h!]
  \centering
  \caption{Ablation study on the RealText-Test set. Both the CCT mechanism and the weighted multi-task reward are shown to be critical for the holistic performance of LogicLens. All metrics are in percentage (\%). Acc: Accuracy, mIOU: mean Intersection over Union, CSS: Cosine Similarity Score, BS(F1): BERTScore F1. M-F1: the macro-average of its Detection F1, Grounding mF1, and Explanation BS(F1). Best results are in \textbf{bold}.}
  \label{tab:ablation}
  \resizebox{\columnwidth}{!}{%
    \begin{tabular}{@{}l|cc|cc|cc|c@{}}
      \toprule
      \multirow{2}{*}{\textbf{Configuration}} & \multicolumn{2}{c|}{\textbf{Detection}} & \multicolumn{2}{c|}{\textbf{Grounding}} & \multicolumn{2}{c|}{\textbf{Explanation}} & \multirow{2}{*}{\shortstack{\textbf{M-F1}}} \\
      \cmidrule(l){2-3} \cmidrule(l){4-5} \cmidrule(l){6-7}
       & Acc & F1 & mIOU & mF1 & CSS & BS(F1) & \\
      \midrule
      \textbf{LogicLens (Full)} & \textbf{91.4} & \textbf{93.2} & \textbf{32.4} & \textbf{36.7} & \textbf{85.0} & \textbf{76.9} & \textbf{68.9} \\
      \midrule
      \textit{w/o} GRPO (SFT only) & 89.4 & 90.3 & 27.6 & 31.1 & 81.9 & 73.2 & 64.9 \\
      \textit{w/o} CCT & 51.7 & 32.6 & 21.5 & 18.2 & 71.3 & 68.1 & 39.6 \\
      \midrule
      \textit{w/o} format reward & 91.2 & 92.6 & 32.1 & 36.0 & 83.8 & 75.4 & 68.0 \\
      \textit{w/o} grounding reward & 90.0 & 91.2 & 28.6 & 32.2 & 84.7 & 76.5 & 66.8 \\
      \textit{w/o} explanation reward & 91.0 & 93.0 & 31.7 & 36.3 & 83.5 & 75.1 & 67.9 \\
      \bottomrule
    \end{tabular}%
  }
\end{table}
As shown in Table~\ref{tab:main_results}, LogicLens significantly outperforms both specialized and general MLLMs. It consistently achieves the highest M-F1 score across RealText, T-IC13, and T-SROIE, establishing a state-of-the-art in unified text-centric forgery analysis. This superiority is evidenced by its exceptional zero-shot generalization on T-IC13 (M-F1 79.8\% vs. 56.4\% for GPT-40) and its remarkable robustness on the challenging T-SROIE dataset. Furthermore, it delivers competitive performance against traditional methods (detailed in Appendix~\textbf{A}).  LogicLens outperforms the specialist DTD on T-IC13 grounding (mF1 67.6\% vs. 45.4\%).

Crucially, our 7B LogicLens model outperforms even larger proprietary systems and its peers, demonstrating that its state-of-the-art performance stems from a superior architecture and training strategy rather than model scale, as visualized in Fig.~\ref{fig:qualitative}. More qualitative comparisons in Appendix~\textbf{C}.
\begin{figure}[t]
  \centering
  \includegraphics[width=\columnwidth]{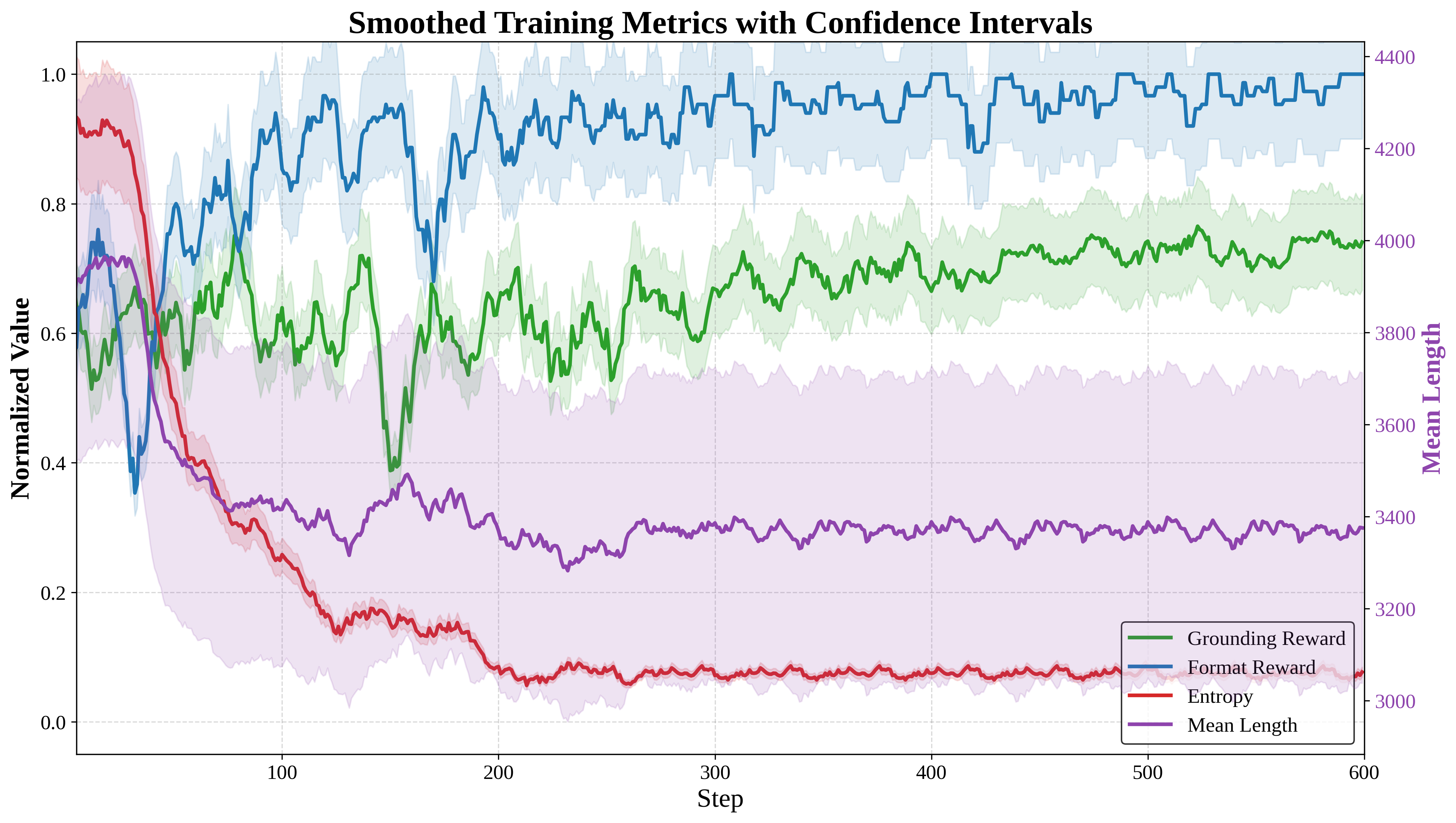}
  \caption{Training curves for the GRPO-based alignment stage, showing smoothed metrics. The simultaneous increase in task-specific rewards (Grounding and Format) alongside a decrease in entropy and mean output length demonstrates the effectiveness of our \textbf{weighted multi-task reward}. This strategy successfully guides the model toward a more stable, confident, and efficient policy, rather than simply a verbose one.}
  \label{fig:training_curves}
\end{figure}
\subsection{Ablation Studies}
\label{subsec:ablation}

\begin{figure*}[t]
  \centering
  \includegraphics[width=\textwidth]{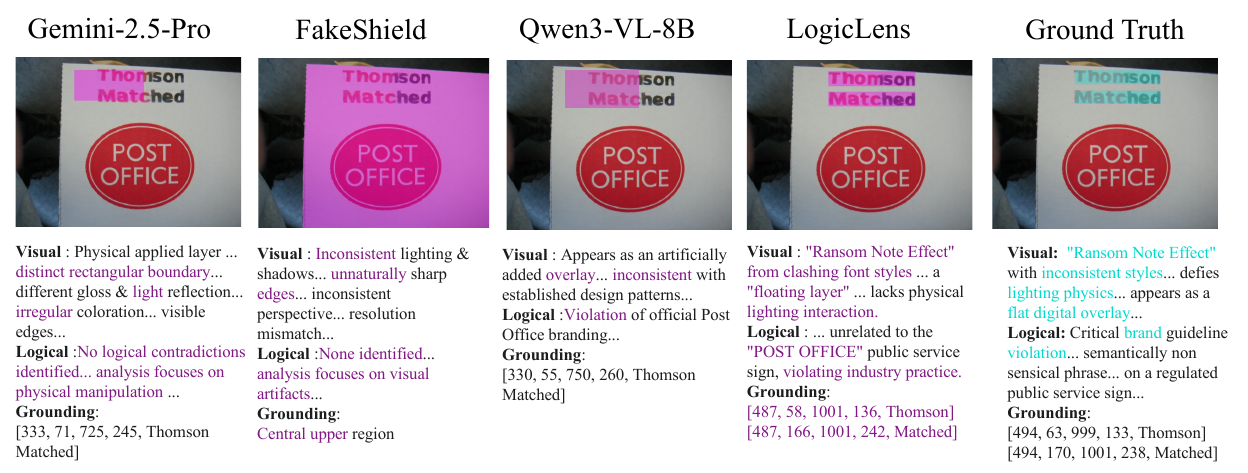}
  \caption{Qualitative comparison of artifact grounding and explanations across different methods. LogicLens demonstrates superior performance, accurately identifying both visual artifacts and logical cues. Shaded regions indicate the localized areas.
  }
  \label{fig:qualitative}
\end{figure*}
Our ablation study, summarized in Table~\ref{tab:ablation}, confirms the critical contributions of both the CCT mechanism and the GRPO alignment stage. The CCT is indispensable for effective reasoning. Its removal results in a severe performance collapse, with the overall M-F1 dropping by 29.3\% and the detection F1 score decreasing by over 60\%. Similarly, the GRPO stage is crucial for refining the model's capabilities, as removing it and relying solely on SFT leads to a significant 4.0\% drop in the overall M-F1 score. These findings confirm that our structured reasoning and reinforcement learning alignment are both essential for LogicLens's performance.

Furthermore, we conducted an in-depth ablation study to investigate the impact of our multi-task reward function. 
As shown in Table~\ref{tab:ablation}, 
the grounding reward proves to be the most critical for discriminative performance; its removal induces the most significant degradation, causing a 2.1\% drop in M-F1 and substantially impairing both detection and grounding. This highlights its central role in driving the model’s ability to identify and localize forgeries.
In contrast, the format adherence and explanation rewards primarily influence the quality of the generated rationale. Specifically, they are essential for producing machine-readable explanations.
The training dynamics of the GRPO stage, shown in Fig.~\ref{fig:training_curves}, illustrate the effectiveness of our optimization strategy. That the model learns the required output structure quickly is demonstrated by the rapid convergence of the Format Reward. 

Crucially, the observation of a significant reduction in entropy indicates that the model is becoming more certain and confident in its predictions.
The mean output length substantially decreases (from approx.~\textbf{4100 to 3400} tokens), implying that the model is learning to be more concise.
This dual reduction is a critical finding, as it confirms that our multi-task reward successfully avoids the rambling outputs simply to maximize scores. 
Instead, it guides the model to achieve superior performance through a more efficient and concise policy. 
This ability to enhance accuracy while simultaneously reducing complexity is the definitive signature of a genuinely successful alignment strategy.

\section{Conclusion}
\label{sec:conclusion}
In this work, we explored the challenging task of text-centric forgery analysis. 
We introduced LogicLens, a unified reasoning framework that reformulates the traditionally separate tasks of detection, grounding, and explanation into a single, joint generative process. 
This framework's deep forgery analysis capabilities are significantly enhanced by our novel Cross-Cues-aware Chain of Thought (CCT) mechanism and a specialized, GRPO-based multi-task reward function. 
To support this approach, we also constructed the RealText dataset, featuring explainable visual and logical anomaly clues, with its multi-task annotations generated via our self-correcting PR² multi-agent pipeline.
Extensive experiments validate our method. LogicLens not only establishes a new state-of-the-art on our RealText benchmark but also demonstrates strong zero-shot generalization and robustness on public datasets. 
These results provide compelling evidence that logical cues are of crucial importance in text-centric forgery analysis, and that visual-textual co-reasoning is essential for this demanding task.
Moreover, the structured textual analysis report produced by LogicLens highlights its potential for real-world impact. 
Its versatility enables numerous downstream applications, from automatically generating professional forensic documents to providing valuable feedback for refining generative models, thus creating a beneficial cycle between generation and detection. 
As this cat-and-mouse game evolves, we hope our public contributions will inspire further research in this critical domain, ultimately bolstering our collective defenses against digital deception.

{
    \small
    \bibliographystyle{ieeenat_fullname}
    \bibliography{main}
}

\appendix

\end{document}